\documentclass{article} 
\usepackage{iclr2025_delta,times}

\usepackage{amsmath,amsfonts,bm}

\def\eqref#1{equation~\ref{#1}}

\def\1{\bm{1}}

\DeclareMathAlphabet{\mathsfit}{\encodingdefault}{\sfdefault}{m}{sl}
\SetMathAlphabet{\mathsfit}{bold}{\encodingdefault}{\sfdefault}{bx}{n}

\usepackage{hyperref}
\usepackage{url}
\usepackage{perl_acronyms}
\usepackage{graphicx}
\usepackage{subfigure}
\usepackage{algorithm}
\usepackage{algorithmicx}
\usepackage{algpseudocode}
\usepackage{adjustbox}
\usepackage{booktabs}
\usepackage{cleveref}

\title{DOSE3 : Diffusion-based Out-of-distribution detection on {$\mathbb{SE}(3)$} trajectories}

\author{Hongzhe Cheng\thanks{Equal Contributor} , Tianyou Zheng$^*$, Tianyi Zhang, Matthew Johnson-Roberson, Weiming Zhi \\
Robotics Institute\\
Carnegie Mellon University\\
Pittsburgh, PA 15213, USA \\
\texttt{\{hongzhec, tianyou2, tianyiz4, mkj, wzhi\}@andrew.cmu.edu} \\
}

\iclrfinalcopy 
\begin{document}

\maketitle

\begin{abstract}
\ac{OOD} detection, a fundamental machine learning task aimed at identifying abnormal samples, traditionally requires model retraining for different inlier distributions. While recent research demonstrates the applicability of diffusion models to \ac{OOD} detection, existing approaches are limited to Euclidean or latent image spaces. Our work extends \ac{OOD} detection to trajectories in the \emph{Special Euclidean Group in 3D} ($\mathbb{SE}(3)$), addressing a critical need in computer vision, robotics, and engineering applications that process object pose sequences in $\mathbb{SE}(3)$. We present \emph{\textbf{D}iffusion-based \textbf{O}ut-of-distribution detection on $\mathbb{SE}(3)$} ($\mathbf{DOSE3}$), a novel \ac{OOD} framework that extends diffusion to a unified sample space of $\mathbb{SE}(3)$ pose sequences. Through extensive validation on multiple benchmark datasets, we demonstrate $\mathbf{DOSE3}$'s superior performance compared to state-of-the-art \ac{OOD} detection frameworks.
\end{abstract}

\section{Introduction}

\ac{OOD} detection represents a fundamental machine learning challenge focused on identifying data samples that deviate from expected inlier distributions. This capability is particularly crucial in safety-critical applications like robotics and autonomous driving, where accurate identification of anomalous motion trajectory \citep{diag_teaching} samples can prevent system failures.
Recent advances in \ac{OOD} detection have explored various unsupervised approaches to learn inlier data representations. These include likelihood-based methods that employ different likelihood measures for \ac{OOD} determination~\citep{NEURIPS2020_8965f766, NEURIPS2019_1e795968, choi2019generative, RAN2022199}, and reconstruction-based approaches that utilize pretrained generative models to assess sample similarity~\citep{denouden2018improvingreconstructionautoencoderoutofdistribution, Wyatt_2022_CVPR, Graham_2023_CVPR}. However, these methods typically require dataset-specific training, necessitating retraining for different \ac{ID} and \ac{OOD} datasets~\citep{heng2024out}. Recent research~\citep{xiao2021reallyneedlearnrepresentations} has addressed this limitation by exploring single discriminative models for \ac{OOD} detection. Our work similarly aims to develop unified OOD approaches that eliminate retraining requirements.

\begin{figure}[t]
    \centering
\fbox{\includegraphics[width=0.75\linewidth]{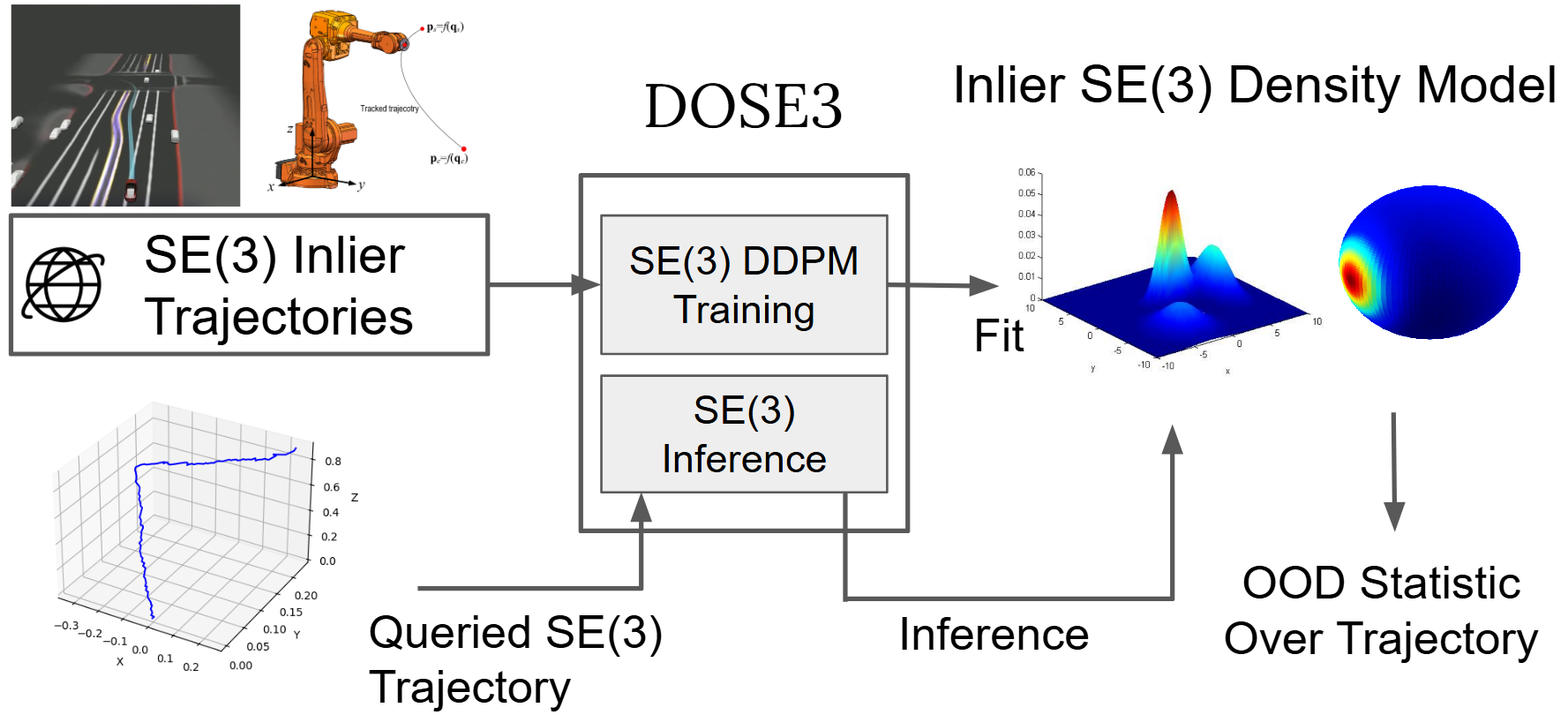}}
    \caption{Sequences of rigid poses are abundant in displines that pertain to objects moving in the real world. We propose $\mathbf{DOSE3}$, a unified diffusion model over the $\mathbb{SE}(3)$ manifold to accurately detect out-of-distribution pose sequences.}
    \label{fig:intro}
\end{figure}

Current trajectory \ac{OOD} detection research primarily focuses on Latent Euclidean spaces, often overlooking explicit manifold space structures. Our work targets OOD detection for rigid body pose data, encompassing both position and orientation information. This type of data is fundamental to numerous applications in physics, engineering, and robotics that analyze object pose evolution over time \citep{unifying, darkgs}. We present theoretical insights and practical algorithms for detecting OOD data in rigid body pose sequences. Our framework, \emph{\textbf{D}iffusion-based \textbf{O}ut-of-distribution detection on $\mathbb{SE}(3)$} ($\mathbf{DOSE3}$), introduces a novel unified generative approach for trajectory space OOD detection. We define a manifold-specific diffusion process for rigid transformations on $\mathbb{SE}(3)$ and develop a high-dimensional OOD statistic for out-of-distribution sample identification.
We validate our approach using established robotics and automation datasets, creating benchmarks from Oxford RobotCar~\citep{RobotCarDatasetIJRR}, KITTI~\citep{Geiger2012CVPR}, and IROS20~\citep{wen2020se}. These datasets enable comprehensive evaluation across varying OOD similarity levels. Our key contributions include:
\begin{enumerate}
\item The $\mathbf{DOSE3}$ framework that \emph{diffuses} over $\mathbb{SE}(3)$ sequences, incorporating manifold structures into OOD detection.
\item A novel OOD statistic derived from our $\mathbb{SE}(3)$ manifold diffusion estimator for sample degree measurement.
\item Comprehensive empirical validation demonstrating $\mathbf{DOSE3}$'s effectiveness in distinguishing between in-distribution and OOD samples across diverse real-world trajectory datasets.
\end{enumerate}
By connecting diffusion models with trajectory OOD detection, $\mathbf{DOSE3}$ advances the development of robust and scalable methods for autonomous systems and 3D trajectory analysis applications.

\section{Related Work}
\textbf{\ac{OOD} detection:}
\ac{OOD} detection plays a crucial role in safety-critical applications such as autonomous driving. Existing methods can generally be categorized into likelihood-based and reconstruction-based approaches. 

Likelihood-based \ac{OOD} detection methods involve training a model on \ac{ID} data and deriving a likelihood statistic from test samples to serve as an \ac{OOD} metric. Early work focused on learning discriminative representations to detect \ac{OOD} samples and identify distributional shifts~\citep{denouden2018improvingreconstructionautoencoderoutofdistribution, NEURIPS2020_8965f766}. More recent research has explored generative models due to their ability to model high-dimensional data and facilitate likelihood estimation~\citep{NEURIPS2020_eddea82a}. However, studies have shown that generative models may assign higher likelihoods to \ac{OOD} samples than to \ac{ID} ones~\citep{nalisnick2019a, hendrycks2019oe}. 

To address this issue, various refinements have been proposed, including likelihood ratios~\citep{NEURIPS2019_1e795968}, \ac{WAIC}~\citep{choi2019generative}, improved noise contrastive priors~\citep{RAN2022199}, and \ac{EBM}s~\citep{liu2020energy}. However, these enhancements remain ineffective in high-dimensional scenarios~\citep{Graham_2023_CVPR}. Another approach considers measuring how \emph{typical} a test input is~\citep{nalisnick2020detecting}, but this method suffers from poor performance at the sample level. Normalizing flows~\citep{NEURIPS2018_d139db6a} have also been investigated for \ac{OOD} detection as they provide direct likelihood estimation, yet they still suffer from overconfidence issues~\citep{NEURIPS2020_ecb9fe2f}. 

Reconstruction-based \ac{OOD} detection methods, on the other hand, aim to reconstruct input samples and compare them to their reconstructions to measure similarity. Early work used the reconstruction probability of VAEs~\citep{an2015variational, kingma2013auto} for anomaly detection. However, later studies found that \ac{OOD} samples can exhibit similar or even lower reconstruction errors compared to \ac{ID} samples, reducing the effectiveness of this approach~\citep{denouden2018improvingreconstructionautoencoderoutofdistribution}. 

\textbf{Diffusion-based OOD Detection:}
Diffusion models (DMs) have achieved remarkable performance in generative tasks across various modalities, including images~\citep{ho2020denoising, song2021scorebased}, videos~\citep{ho2022video}, and audio~\citep{chen2020wavegradestimatinggradientswaveform}. More recently, research has emphasized the robustness of DMs in sampling and their potential use in \ac{OOD} detection. Utilizing the reconstruction mean squared error (MSE) of DDPMs as an \ac{OOD} score has been shown to enhance image-space \ac{OOD} detection~\citep{Wyatt_2022_CVPR, Graham_2023_CVPR}. However, these models require retraining for different in-distribution datasets. 

A growing trend in machine learning research is the development of unified learning frameworks that generalize across various tasks~\citep{xiao2021reallyneedlearnrepresentations}. In an effort to construct a unified DM for \ac{OOD} detection, DiffPath~\citep{heng2024out} demonstrated that the rate of change and curvature of the forward diffusion trajectory can serve as effective \ac{OOD} metrics, eliminating the need for retraining on different datasets. 

The aforementioned \ac{OOD} detection methods—whether likelihood-based or reconstruction-based—are primarily focused on images or Euclidean space data. In contrast, some research from the robotics community implicitly incorporates \ac{OOD} detection under the framework of trajectory planning or optimization in 2D spaces~\citep{PDMP, 5354448, 5509799}. However, these models struggle to generalize to complex real-world scenarios that involve three-dimensional interactions~\citep{wang2023efficienttrajectorygenerationground, 10.1016/j.vehcom.2024.100733}. While there is a growing interest in extending diffusion models to non-Euclidean spaces ~\citep{huang2022riemannian, leach2022denoising}, these methods are limited to generating individual samples on manifolds rather than modeling entire trajectories. 

To the best of our knowledge, \textbf{DOSE3} is the first approach to leverage manifold-based diffusion over entire trajectories, enabling a unified \ac{OOD} detection framework.

\section{Preliminaries}
In this section, we first provide background on the architecture of diffusion models. We then discuss the recent advancements in constructing \emph{Unified} OOD detection models using diffusion models. Finally, we introduce the \emph{Special Euclidean Group in 3D}, $\mathbb{SE}(3)$, and elaborate on its geometric structure and related statistical foundations.

\subsection{Denoising Diffusion Probabilistic Model (DDPM)}

Diffusion models have gained widespread attention in generative modeling due to their strong ability to synthesize high-fidelity data. These models employ a forward diffusion process, where data $\mathbf{x_0}$ is gradually corrupted by adding Gaussian noise over $T$ timesteps, ultimately producing a noisy distribution $\mathbf{x_T}$ that approximates a standard normal distribution. The goal is to learn the reverse diffusion process, which systematically denoises $\mathbf{x_T}$ to recover the original data distribution. 

At the core of this reverse process is the $\epsilon$-model, typically implemented as a neural network trained to predict the noise $\boldsymbol{\epsilon}$ added at each timestep $t$. The forward diffusion process, expressed in equation~\ref{eq:forward}, illustrates how standard Gaussian noise is introduced to perturb the original sample $\mathbf{x}_0$. The backward process, given in equation~\ref{eq:backward}, employs the estimator model $\boldsymbol{\epsilon}_\theta$, which estimates the true Gaussian noise $\boldsymbol{\epsilon}$ and enables data recovery by removing the noise.

\begin{align}
\mathbf{x}_t &= \sqrt{\bar{\alpha}_t} \mathbf{x}_0 + \sqrt{1 - \bar{\alpha}_t} \boldsymbol{\epsilon}, \quad \boldsymbol{\epsilon} \sim \mathcal{N}(\mathbf{0}, \mathbf{I}) \label{eq:forward} \\
\mathbf{x}_{t-1} &= \frac{1}{\sqrt{\alpha_t}} \left(\mathbf{x}_t - \frac{\beta_t}{\sqrt{1 - \bar{\alpha}_t}} \boldsymbol{\epsilon}_\theta(\mathbf{x}_t, t) \right) + \sigma_t \mathbf{z} \nonumber\\
&\quad \mathbf{z} \sim \mathcal{N}(\mathbf{0}, \mathbf{I}) \label{eq:backward}
\end{align}
where $\alpha_t$, $\beta_t$, and $\bar{\alpha}_t$ are predefined noise schedule parameters, and $\mathbf{z} \sim \mathcal{N}(0, \mathbf{I})$. 

The theoretical foundation of diffusion models is grounded in variational inference, where the evidence lower bound (ELBO) in equation~\ref{eq:ELBO} is maximized to ensure that the learned reverse process closely approximates the true data distribution.
\begin{align}
\mathcal{L}_{\text{ELBO}} = \mathbb{E}_{q} [ & D_{\text{KL}}( q(\mathbf{x}_T | \mathbf{x}_0) \parallel p(\mathbf{x}_T) ) 
+ \sum_{t=2}^{T} D_{\text{KL}}( q(\mathbf{x}_{t-1} | \mathbf{x}_t, \mathbf{x}_0) \parallel p_\theta(\mathbf{x}_{t-1} | \mathbf{x}_t) ) - \log p_\theta(\mathbf{x}_0 | \mathbf{x}_1) ]
\label{eq:ELBO}
\end{align}
By leveraging the $\epsilon$-model within this framework, diffusion models effectively capture complex data manifolds, achieving state-of-the-art generative performance.

\subsection{Unified Out-of-Distribution Detection}
Traditional \ac{OOD} detection methods, such as likelihood-based and reconstruction-based approaches, require retraining a new model for each specific inlier data distribution. This results in significant computational costs when switching between different OOD tasks and distributions. Recently, \cite{heng2024out} introduced a new concept of Unified \ac{OOD} detection, where a single unconditional diffusion model is trained, and distributional information can be obtained from inlier distributions that were unseen during training.

The theoretical foundation of this approach builds on the variance-preserving formulation used in DDPM. The difference between each denoising timestep is given in equation~\ref{eq:diffusion_process} and can be rewritten as:
\begin{align}
    d\mathbf{x}_t &= -\frac{1}{2} \beta_t \mathbf{x}_t \, dt + \sqrt{\beta_t} \, d\mathbf{w}_t, \quad \mathbf{x}_0 \sim p_0(\mathbf{x}) \label{eq:diffusion_process} \\
    \frac{d \mathbf{x}_t}{dt} &= f(\mathbf{x}_t, t) + \frac{g(t)^2}{2 \sigma_t^2} \epsilon_p(\mathbf{x}_t, t) \label{eq:sgm_rewrite}
\end{align}
In equation~\ref{eq:kl_divergence}, we denote $\phi_T$ and $\psi_T$ as the marginals obtained by evolving two distinct distributions, $\phi_0$ and $\psi_0$, using their respective probability flow ordinary differential equations (ODEs) from equation~\ref{eq:sgm_rewrite}.
\begin{align}
    D_{\mathrm{KL}} (\phi_0 \parallel \psi_0) = \frac{1}{2} \int_{0}^{T} \mathbb{E}_{\mathbf{x}_t \sim \phi_t} \left[ \frac{g(t)^2}{\sigma_t^2} \left\| \epsilon_{\phi}(\mathbf{x}_t, t) - \epsilon_{\psi}(\mathbf{x}_t, t) \right\|^2 \right] dt + D_{\mathrm{KL}}(\phi_T \parallel \psi_T)\label{eq:kl_divergence}
\end{align}
However, the KL divergence remains dependent on the specific model estimators $\epsilon_\phi$ and $\epsilon_\psi$ in equation~\ref{eq:kl_divergence}. The key observation is that even when executing DDPM forward diffusion using an estimator $\epsilon_\theta$ trained on a third distribution $\theta$, the sample can still be successfully transformed into a standard Gaussian distribution. This insight motivates the use of $\epsilon_\theta$—metrics extracted from an arbitrary diffusion estimator—to perform OOD detection on an inlier distribution $\phi$.

\begin{figure*}[t]
    \centering
\includegraphics[width=0.98\linewidth]{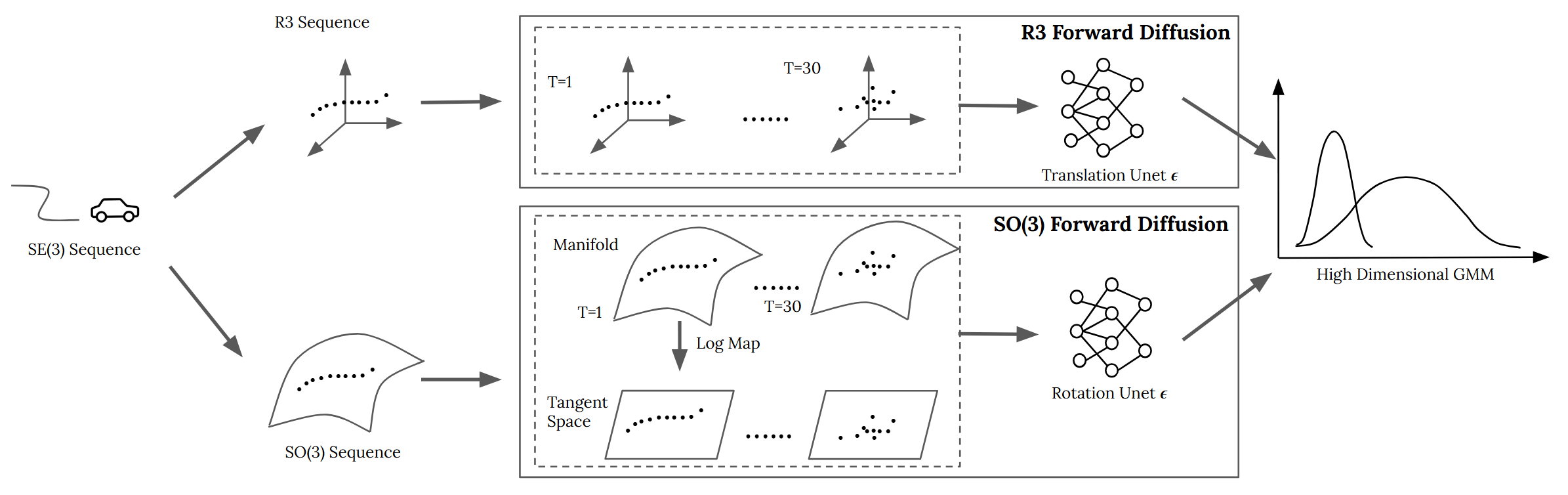}
    \caption{System Diagram of DOSE3 processing flow. Sequences of pose data are diffused, where diffusion over rotational components is constrained to the $\mathbb{SO}(3)$ manifold. The resulting diffusion estimators are used to construct an OOD statistic.}
    \label{fig:system}
\end{figure*}

\subsection{The Special Euclidean Group in 3D} \label{subsec:SE(3)}

The Special Euclidean Group in 3D, denoted as \( \mathbb{SE}(3) \), represents the space of rigid body transformations, which consist of both rotations and translations. The transformation can be written as:
\[
T = \begin{bmatrix}
R & t \\
0 & 1
\end{bmatrix}
\]
where \( R \in \mathbb{SO}(3) \) is a rotation matrix, and \( t \in \mathbb{R}^{3} \) is a translation vector. The group \( \mathbb{SO}(3) \) consists of all \( 3 \times 3 \) real orthogonal matrices with determinant equal to one:
\begin{equation}
    \mathbb{SO}(3) = \{ R \in \mathbb{R}^{3 \times 3} \mid R^\top R = I, \ \det(R) = 1 \}
\end{equation}
where \( I \) is the \( 3 \times 3 \) identity matrix. The group \( \mathbb{SO}(3) \) represents all possible rotations about the origin in three-dimensional space.

The Lie algebra associated with \( \mathbb{SO}(3) \) is denoted as \( \mathfrak{so}(3) \) and consists of all \( 3 \times 3 \) skew-symmetric matrices. A general element \( \Omega \in \mathfrak{so}(3) \) can be written as:
\[
\Omega = \begin{bmatrix}
0 & -\omega_3 & \omega_2 \\
\omega_3 & 0 & -\omega_1 \\
-\omega_2 & \omega_1 & 0
\end{bmatrix}
\]
where \( \omega = [\omega_1, \omega_2, \omega_3]^\top \) is a vector in \( \mathbb{R}^3 \). The Lie algebra \( \mathfrak{so}(3) \) serves as the tangent space to the manifold \( \mathbb{SO}(3) \), providing a locally Euclidean structure that facilitates computations on \( \mathbb{SO}(3) \).

The exponential map, \( \exp: \mathfrak{so}(3) \rightarrow \mathbb{SO}(3) \), maps an element from the Lie algebra to the Lie group, enabling the representation of rotations in matrix form. Given \( \Omega \in \mathfrak{so}(3) \), the exponential map is defined as:
\begin{equation}
    \exp(\Omega) = I + \frac{\sin \theta}{\theta} \Omega + \frac{1 - \cos \theta}{\theta^2} \Omega^2
    \label{eq:exp}
\end{equation}
where \( \theta = \|\omega\| \) is the rotation angle, and \( \omega \) is the vector corresponding to \( \Omega \).

Conversely, the logarithmic map, \( \log: \mathbb{SO}(3) \rightarrow \mathfrak{so}(3) \), converts a rotation matrix into its corresponding Lie algebra representation. For any \( R \in \mathbb{SO}(3) \) that is not the identity matrix, the logarithmic map is given by:
\begin{equation}
    \log(R) = \frac{\theta}{2 \sin \theta} (R - R^\top)
    \label{eq:log}
\end{equation}
where the rotation angle \( \theta \) is computed as:
\begin{equation}
    \theta = \cos^{-1}\left( \frac{\text{trace}(R) - 1}{2} \right)
\end{equation}
Here, \( \mathfrak{so}(3) \), the \emph{tangent space} of \( \mathbb{SO}(3) \), lies within Euclidean space, allowing standard algebraic operations to be applied. This property is particularly useful for designing diffusion models over \( \mathbb{SO}(3) \), as it enables efficient computations and parameterizations of rotations.

\section{Method}
\subsection{Overview}
Here, we present \emph{\textbf{D}iffusion-based \textbf{O}ut-of-distribution detection on $\mathbb{SE}(3)$}, $\mathbf{DOSE3}$. $\mathbf{DOSE3}$ introduces a \emph{unified} diffusion model for rigid pose trajectories, specifically designed to accommodate the $\mathbb{SE}(3)$ manifold structure.
We first detail $\mathbf{DOSE3}$'s model architecture for handling ordered sequences. We then introduce \emph{$\mathbb{SE}(3)$ Denoising Diffusion Probabilistic Models} ($\mathbb{SE}(3)$ - DDPM), outlining their training and inference algorithms that incorporate rigid pose structure into the diffusion model. Finally, we explain how to utilize the \emph{diffusion estimator}, a function naturally emerging from $\mathbb{SE}(3)$ - DDPM, to develop an OOD detection statistic for evaluating test samples.
\subsection{Architectural details of DOSE3}
\begin{figure}[t]
    \centering
\includegraphics[width=0.98\linewidth]{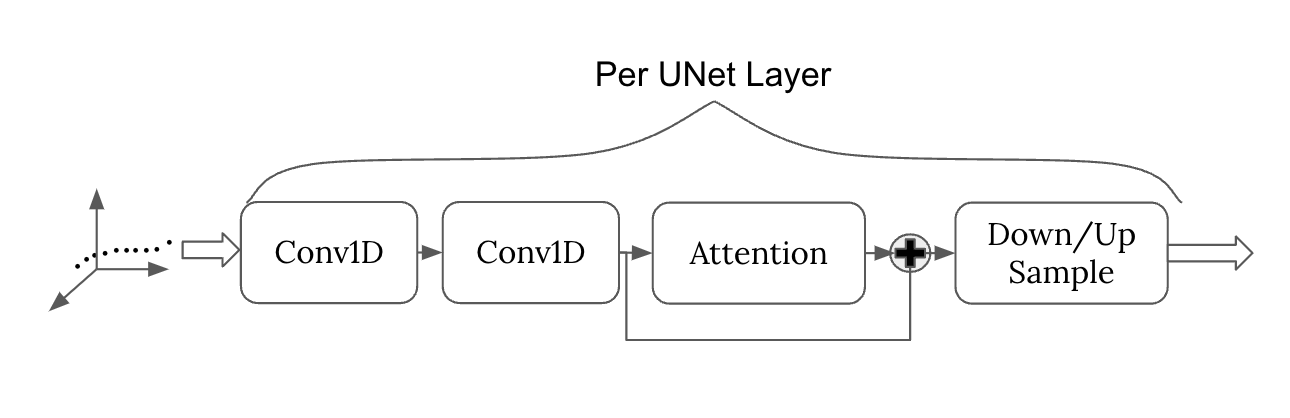}
    \caption{SE(3) diffusion UNet architecture per layer. Trajectories are processed by the 1-dimensional convolution modules.}
    \label{fig:unet}
\end{figure}

The \textbf{UNet} architecture, widely adopted in diffusion models for its effective encoder-decoder structure, enables high-fidelity data generation. Originally developed for biomedical image segmentation, UNet's symmetric design with skip connections preserves spatial information through its network layers. While the original UNet employs 2D convolution layers with max pooling and up convolution for dimensional adjustment, we modify this architecture for sequential data diffusion through the following enhancements:
\begin{enumerate}
\item Replace all convolution layers with 1D convolutions to process temporal structures in motion trajectories.
\item Introduce attention layers before each up- or down-sampling operation to better capture long-range dependencies in trajectory data, extending beyond the local computations of convolution layers.
\item Implement Residual connections~\citep{resnet} around attention layers, similar to Transformer architecture, to enhance learning capabilities for our complex data format and task.
\end{enumerate}
The resulting architecture for each up/down UNet layer in our model is illustrated in \cref{fig:unet}.

\subsection{Training and Inference of $\mathbf{DOSE3}$}
As discussed in section~\ref{subsec:SE(3)}, the incorporation of rotation matrices from $\mathbb{SE}(3)$ format introduces manifold space considerations that preclude direct application of classical diffusion algorithms. We address three primary challenges:
(1) The undefined nature of addition and scalar multiplication operations for rotation matrices;
(2) The inability to guarantee valid rotation matrices when sampling $3\times 3$ matrices from $\mathcal{N}(0,I)$;
(3) The inadequacy of simple $L2$ norm differences for measuring distances/losses between rotation matrices.
To overcome these challenges, we introduce the new $\mathbb{SE}(3)$ DDPM algorithm. While we apply standard Euclidean space diffusion to the translational components of $\mathbb{SE}(3)$, we develop specialized techniques for handling manifold diffusion over the $\mathbb{SO}(3)$ rotation space.

We redefine the operators $\in$ $\mathbb{SO}(3)$ as follows. Essentially, we perform all operations after transforming the $\mathbb{SO}(3)$ data from manifold space into Euclidean tangent space by exponential and logarithmic map given by \cref{eq:exp} and \cref{eq:log}.
\begin{align}
    R_1 \oplus R_2 &= R_1 \cdot R_2\\
    k \otimes R_1 &= \exp(k \cdot \log(R_1)) \\
    k \in \mathbb{R},\  &R_1, R_2 \in \mathbb{SO}(3) \nonumber
\end{align}
We then change the noise sampling method from standard Gaussian distribution to the Isotropic Gaussian distribution on $\mathbb{SO}(3)$ ( $\mathcal{IG}_{\mathbb{SO}(3)}$ ) distribution. Shown in equation~\ref{eq:igso3}, we first sample v from standard Gaussiance distribution, representing the tangent vector, and then use the exponential map operation to transform it to the $\mathbb{SO}(3)$ space.
\begin{equation}
    \mathcal{IG}_{\mathbb{SO}(3)}(\mu, \sigma^2) = \mu \otimes v, \quad v \in \mathbb{R}^3 \sim \mathcal{N}(0,\sigma^2 I)
    \label{eq:igso3}
\end{equation}

Combining all the metrics and operations defined above, we design the full forward and backward $\mathbb{SO}(3)$ DDPM equations, incorporating the operations on the manifold, as,
\begin{align}
q(x_t \mid x_0) &= \bigl(\sqrt{\bar{\alpha}_t} \otimes x_0\bigr) \oplus \bigl((1 - \bar{\alpha}_t) \otimes \epsilon \bigr), \text{where}\  \epsilon\!\sim\!\mathcal{IG}_{\mathbb{SO}(3)}(0, I)\nonumber\\
\widehat{x}_0 \!&= \!\frac{1}{\sqrt{\bar{\alpha}_t}} \!\otimes\!\left(x_t \oplus \left(- \sqrt{1 - \bar{\alpha}_t} \!\otimes \!\epsilon_\theta(x_t, t)\right)\right), \\
\mu_t &= \left(\frac{\sqrt{\bar{\alpha}_{t-1}} \beta_t}{1 - \bar{\alpha}_t} \otimes \widehat{x}_0\right) \oplus \nonumber  \left(\frac{\sqrt{\alpha_t (1 - \bar{\alpha}_{t-1})}}{1 - \bar{\alpha}_t} \otimes x_t\right), \\
x_{t-1} &=\!\mu_t\!\oplus\!\left(\!\sqrt{\beta_t} \!\otimes \epsilon\right)
\end{align}
The rotational distance will be adapted as the loss for training in $\mathbb{SO}(3)$ . The metric will reflect the average angle difference on each axis for two rotation matrices. The equation can be written as
$$\text{L}_{rot}(R_1, R_2) = \arccos\left(\frac{\mathrm{trace}(R_1^{\top} R_2) - 1}{2}\right)^2$$ 
The final $\mathbb{SE}(3)$ diffusion training takes in batches of trajectories in the format of ordered $\mathbb{SE}(3)$ sequences, maintaining the ordering of the trajectories. The complete training pseudo-code is shown in Algorithm~\ref{algo:SE3_diffusion}.

\begin{algorithm}[t]
\caption{SE(3) Sequence DDPM Training}
\label{algo:SE3_diffusion}
\begin{algorithmic}[1]
    \Require \text{Estimator}-$\epsilon_\theta$,\text{ Diffusion Scheduler}-$\alpha, \bar\alpha$, \text{ Train $\mathbb{SE}(3)$ Dataset =[trans, rot]}
    \While{training}
        \State $z_{\text{rot}} \gets \mathcal{IG}_{SO(3)}(0, I)$
        \State $z_{\text{trans}} \gets \mathcal{N}(0, I)$
        \State $(\text{trans}_t, \text{rot}_t) \gets \text{diffusion}(\text{rot}, \text{trans}, z_{\text{rot}}, z_{\text{trans}}, t)$
        \State $(\text{score}_{\text{trans}}, \text{score}_{\text{rot}}) \gets \text{model}(x_{\text{t-rot}}, x_{\text{t-trans}}, t)$
        \State $\text{trans}_0 \gets \frac{\text{trans} - \sqrt{1 - \bar{\alpha}_t} \cdot \text{score}_{\text{trans}}}{\sqrt{\bar{\alpha}_t}}$
        \State $\text{rot}_0 \gets \exp\left(\frac{\log(\text{rot}) - \sqrt{1 - \bar{\alpha}_t} \cdot \text{score}_{\text{rot}}}{\sqrt{\bar{\alpha}_t}}\right)$
        \State $\text{loss}_{\text{x0}} \gets \text{L1}(\text{trans}_0, \text{trans}) + \text{L}_{rot}(\text{rot}_0, \text{rot})$
        \State $\text{loss}_{\epsilon} \gets \text{L1}(\text{score}_{\text{trans}}, z_{\text{trans}}) + \text{L}_{rot}(\text{score}_{\text{rot}}, z_{\text{rot}})$
        \State $\text{loss} \gets \text{loss}_{\epsilon} + \text{loss}_{\text{x0}}$
        \State $\text{loss.backward()}$
    \EndWhile
\end{algorithmic}
\end{algorithm}

\subsection{OOD Detection}
\begin{figure}[t]
    \centering
    \subfigure[Oxford]{
        \includegraphics[width=0.46\linewidth]{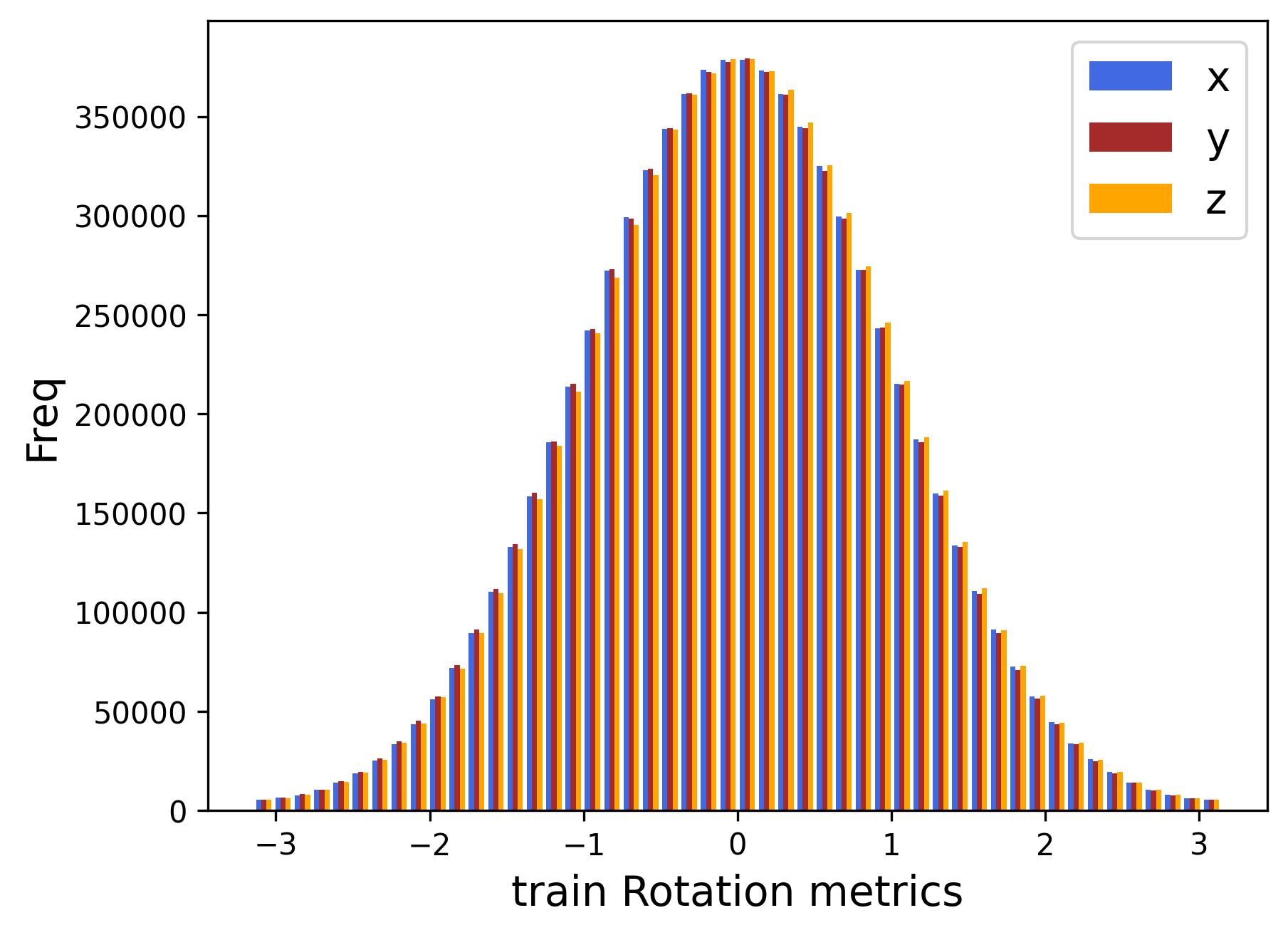}
    }%
    \subfigure[KITTI]{
        \includegraphics[width=0.46\linewidth]{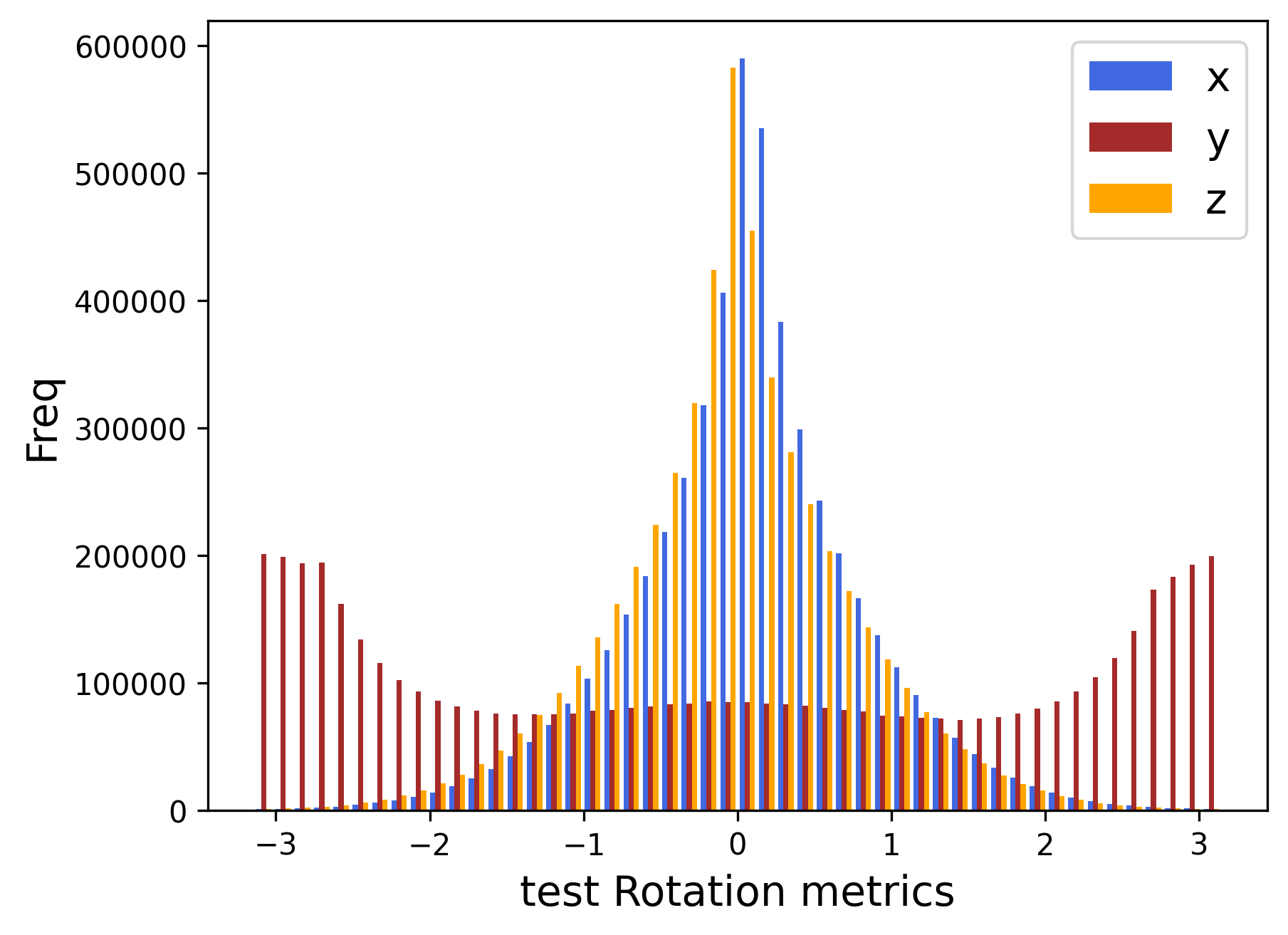}
    }
    \caption{Distribution of elements $\epsilon(x_t, t)$ in rotation tangent space for each dimension when running Oxford Robot Car and KITTI dataset on model trained on Oxford Robot Car}
    \label{fig:rot_3d}
\end{figure}

While likelihood-based \ac{OOD} detection algorithms traditionally rely on generative model likelihood measures, the ELBO shown in equation~\ref{eq:ELBO} has proven inadequate for \ac{OOD} tasks due to its tendency to overestimate OOD sample likelihood~\citep{Serrà2020Input}. Recent research demonstrates that the diffusion estimator $\epsilon_\theta$ and its derivatives effectively capture data distribution characteristics and can be obtained from a unified diffusion model without retraining. As shown in Equation~\ref{eq:kl_divergence}, the norm of noise estimator $\epsilon$ correlates with the divergence between different data distributions~\citep{heng2024out}. Based on this insight, we define the following OOD statistics group for a diffusion model with noise estimator $\epsilon_\theta$, where the operator $\langle x\rangle_p = \frac{1}{N}\sum_{i=0}^N x_i^p$.
\begin{align}
    Me&tricGroup(\epsilon_{\theta}) = \nonumber \\
    \Bigg[&\sum_t\langle\epsilon_{\theta}(x_t, t)\rangle_1, \sum_t\langle\epsilon_{\theta}(x_t, t)\rangle_2, \sum_t\langle\epsilon_{\theta}(x_t, t)\rangle_3 \nonumber\\
    &\sum_t\langle\partial\epsilon_{\theta}(x_t, t)\rangle_1, \sum_t\langle\partial\epsilon_{\theta}(x_t, t)\rangle_2, \sum_t\langle\partial\epsilon_{\theta}(x_t, t)\rangle_3 \Bigg],
\end{align}
where $MetricGroup(\epsilon_{\theta})\in\mathbb{R}^{6}$. For each sample $x_0$, we apply the DDPM forward process to obtain the perturbed sample $x_t$, then compute the metric group to derive final statistics. Given that our $\mathbb{SE}(3)$ diffusion model comprises separate sub-diffusions for $\mathbb{R}^{3}$ and $\mathbb{SO}(3)$, and rotation metric distributions can vary along x, y, and z directions (as illustrated in figure~\ref{fig:rot_3d}), we establish distinct metric sets for each rotational dimension. This results in 4 sets of statistics per sample: three for individual rotational axes and one for translation, yielding a total of 24 metrics per sample.
To process these metrics from the inlier data distribution, we estimate the density over the $24$-dimensional joint vector of metric groups. We employ straightforward density estimators such as Gaussian Mixture Models or Kernel Density Estimators, as each metric empirically exhibits approximately Gaussian behavior. During testing, we collect identical metrics for each test sample and infer likelihood from the inlier density estimator, identifying lower-likelihood samples as out of distribution. We establish the \ac{OOD} sample threshold at the bottom 5 percentile of inlier distribution likelihoods. The complete OOD model fitting and inference procedure is detailed in Algorithm~\ref{algo:OOD}.

\begin{algorithm}[t]
\caption{OOD Detection}
\label{algo:OOD}
\begin{algorithmic}[1]
\Require \text{DDPM Model, Inlier $\mathbb{SE}(3)$ Dataset I}, \text{Query $\mathbb{SE}(3)$ Trajectory q}
\State $stats \gets []$
\For{$traj_0 \text{ in I}$}
\State $sum \gets [0,0,0,0,0,0]$
\For{$\text{t in [0, T-1]}$}
    \State $traj_t \gets DDPM_{forward}(traj_0, t)$
    \State $sum \gets sum + \text{6D metrics over }\epsilon_\theta(traj_t, t)$
\EndFor
\State $stats.\text{append}(sum)$
\EndFor
\State $distribution \gets \text{GMM.fit(stats)}$
\State $q_t \gets DDPM_{forward}(q, t)$
\State $metric_q \gets \text{6D metrics over }\epsilon_\theta(q_t, t))$
\State $likelihood \gets distribution\text{.eval(}metric_q\text{)}$
\State $OOD \gets likelihood < threshold$
\end{algorithmic}
\end{algorithm}

\section{Experiments}
\subsection{Experiment Setup}
To evaluate $\mathbf{DOSE3}$'s validity, performance, and comprehensiveness, we conduct \ac{OOD} testing using the following $\mathbb{SE}(3)$ datasets:
\begin{itemize}
\item $\textbf{Oxford RobotCar}$~\citep{RobotCarDatasetIJRR}: This autonomous driving dataset encompasses over 1000 km of driving data from central Oxford, UK. It features multiple sensor modalities, including high-resolution stereo and monocular cameras, 2D and 3D LiDAR scans, and GPS/INS ground truth localization. Our experiments utilize the 3D LiDAR scans and ground-truth poses stored in $\mathbb{SE}(3)$ format.
\item $\textbf{KITTI}$~\citep{Geiger2012CVPR}: This comprehensive odometry dataset captures autonomous driving scenarios across urban, suburban, and rural environments. The dataset provides stereo and monocular camera imagery, 3D point clouds from a Velodyne LiDAR, and precise GPS/INS measurements. We utilize its pose data represented in $\mathbb{SE}(3)$
\item $\textbf{iros20-6d-pose-tracking}$~\citep{wen2020se}: This dataset advances research in 6D object pose estimation and tracking in dynamic environments. It is specifically designed to support the development and evaluation of algorithms for accurately determining and tracking six degrees of freedom (6D) poses in real-world scenarios.
\end{itemize}
For our diffusion model implementation, we standardize the input length for both $\mathbb{R}^{3}$ and $\mathbb{SE}(3)$ trajectory diffusion. Each trajectory in the datasets is segmented into fixed-length sub-paths of size 128 during experiments.
To standardize the translation data, we first center each trajectory by setting its starting coordinate to the origin, then normalize by dividing by the maximum translation value. This process constrains the translation data to the range [-1, 1], ensuring the model learns trajectory geometry independent of scale.
We evaluate $\mathbf{DOSE3}$ against leading \ac{OOD} detection methods, including \ac{JEM}\citep{Grathwohl2020Your} and Glow Model\citep{NEURIPS2018_d139db6a} with Likelihood Ratio~\citep{NEURIPS2019_1e795968}. These established baselines effectively handle high-dimensional inputs and are widely used for \ac{OOD} detection in image datasets.
\subsection{Quantitative evaluation of $\epsilon_{\theta}$ Distribution as an OOD Metric}
\begin{figure}[h!]
\centering
    \subfigure[Translation Metric]{
        \includegraphics[width=0.45\linewidth]{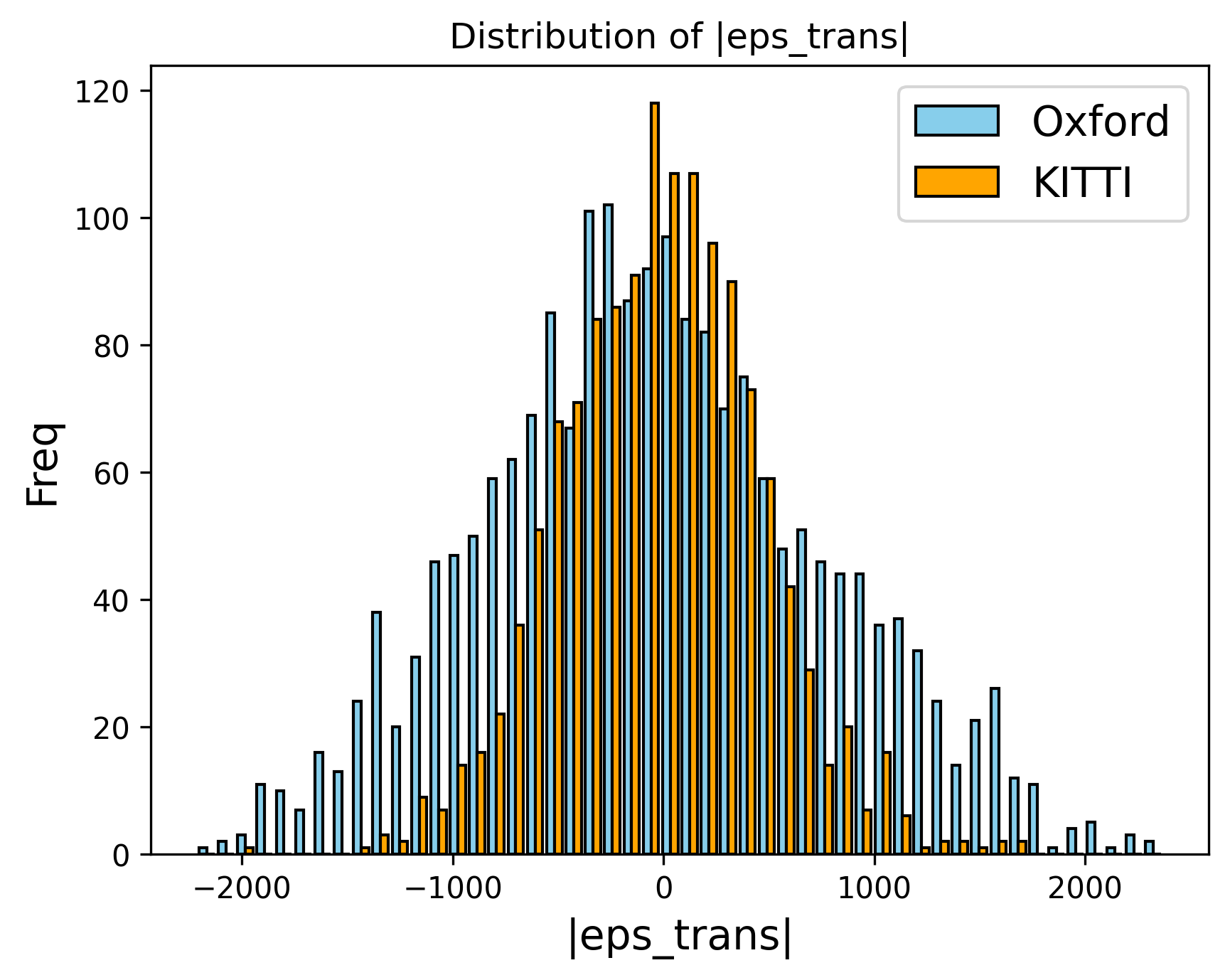}
    }
    \hfill
    \subfigure[Rotation Z axis metric]{
        \includegraphics[width=0.45\linewidth]{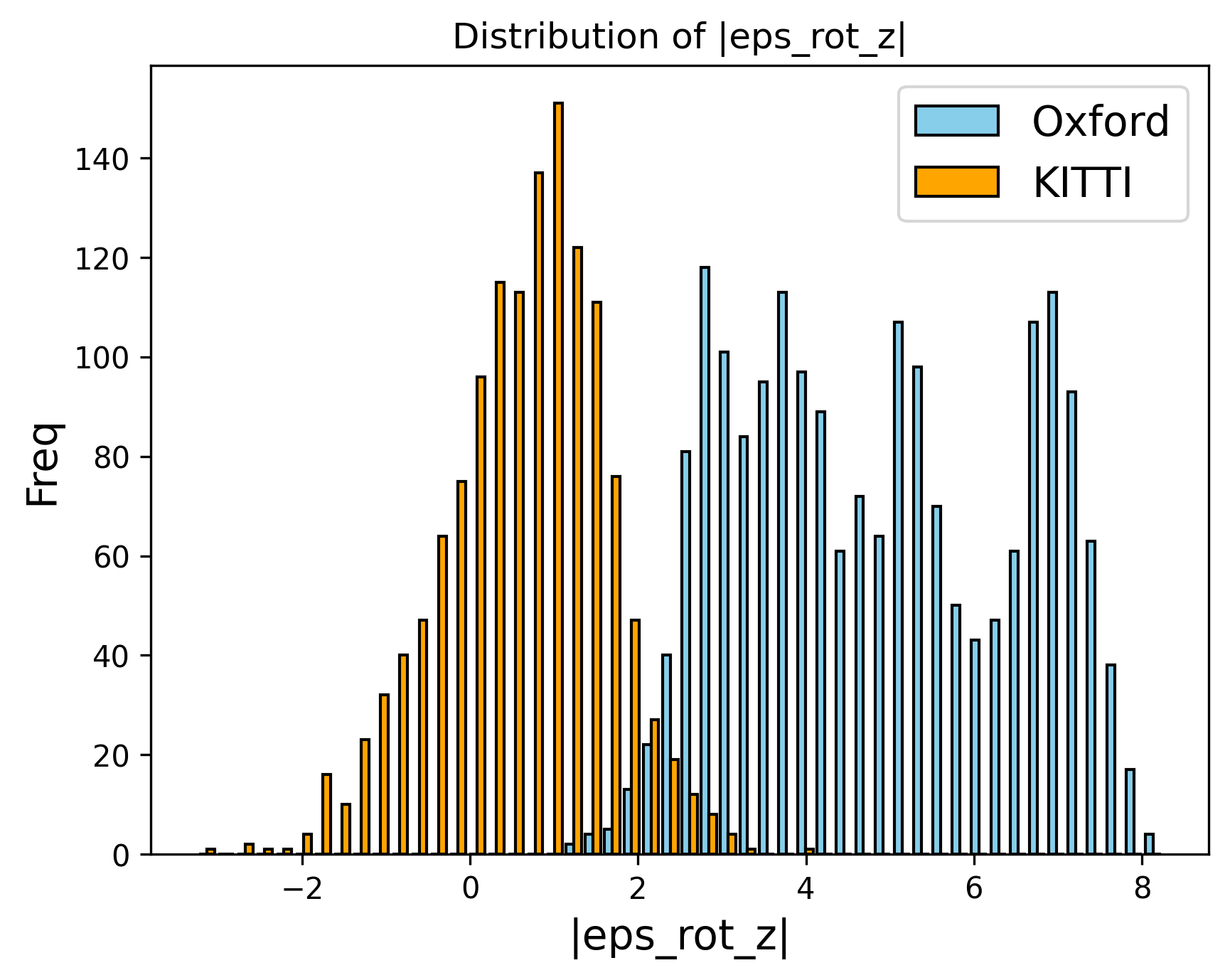}
    }
\vskip -0.2in
\caption{$\epsilon$ distribution retrieved by OOD testing on a model trained from KITTI dataset against the Oxford RoboCar dataset. We observe that the distributions of both datasets is particularly evident on the Z-axis rotation.}
\label{fig:eps_oxford_kitti}
\end{figure}

We analyze the statistical distribution of $\epsilon_{\theta}$ from inlier data to assess its effectiveness as an \ac{OOD} detection metric. Specifically, we investigate how the $\epsilon_{\theta}$ distribution of the $\mathbb{SO}(3)$ diffusion contributes to OOD sample identification. In~\cref{fig:eps_oxford_kitti}, we present a comparative analysis of $\epsilon$ distributions between Oxford RobotCar and KITTI datasets, using a model trained on KITTI. Our findings reveal that after translation data normalization, the translation $\epsilon$ distributions show substantial overlap across datasets, making them unsuitable as reliable OOD indicators. However, the rotation distribution, especially along the z-axis, demonstrates clear dataset separation. For the KITTI-trained model, we observe that KITTI's rotation distribution is centered at 0, aligning with standard Gaussian noise sampling characteristics. In contrast, the Oxford RobotCar dataset exhibits a notable rightward shift in its distribution, suggesting that reconstructing a KITTI sample from Oxford RobotCar input requires a non-Gaussian sampling distribution.

\subsection{Quantitative Results}
Table~\ref{tab:results} presents the OOD detection performance across datasets using the AUROC metric. All evaluated models underwent unsupervised training exclusively on the KITTI dataset. Our $\mathbb{SE}(3)$ model demonstrates exceptional performance, achieving near-perfect AUROC scores across all \ac{ID} and \ac{OOD} dataset combinations. In contrast, JEM, Glow-LR, and the $\mathbb{R}^{3}$ model show degraded performance when evaluating KITTI as an \ac{OOD} dataset, or when testing on the previously unseen Oxford Robot and IROS20 datasets. Additionally, we evaluate the impact of rotation metric splitting in $\mathbb{SE}(3)$ diffusion. The results indicate that separating the rotation metric space substantially enhances $\mathbf{DOSE3}$'s robustness and unified feature representation, particularly improving AUROC scores in scenarios where the training dataset, KITTI, serves as the \ac{OOD} data.
\begin{table}[t]
\caption{AUROC $\uparrow$ of OOD Detection. $\textbf{Bold}$ denotes the best
result.\\(O: Oxford, K: KITTI, I:IROS20)}
    \vskip 0.15in
    \begin{adjustbox}{width=0.68\linewidth,center}
    \begin{tabular}{lccccccc}
    \toprule
        \textbf{Method} & \textbf{O/K} & \textbf{K/O} & \textbf{O/I} & \textbf{I/O} & \textbf{K/I} & \textbf{I/K}\\
    \midrule
        JEM & 0.211 & 0.786 & 0.437 & 0.561 & 0.631 & 0.336\\
        Glow-LR  & 0.461 & 0.556 & 0.470 & 0.539 & 0.529 & 0.454\\
    \midrule
        $\mathbb{R}^{3}$-KITTI  & 0.362 & 0.770 & 0.417 & 0.585 & 0.793 & 0.409\\
        SE3-KITTI  & 0.845 & 0.952 & \textbf{1.000} & 0.398 & \textbf{1.000} & 0.234\\
          (No split) &&&&&&\\
        SE3-KITTI  & \textbf{1.000} & \textbf{0.956} & \textbf{1.000} & \textbf{0.931} & \textbf{1.000} & \textbf{0.897}\\
    \bottomrule
    \end{tabular}
    \end{adjustbox}
    \vskip -0.1in
    \label{tab:results}
\end{table}

\subsection{Ablations}
\subsubsection{Trajectory Dataset used for Training}
$\mathbf{DOSE3}$ strives to develop a single unified model for effective OOD detection. We evaluate both $\mathbb{R}^3$ and $\mathbb{SE}(3)$-based diffusion models trained on different datasets. Table~\ref{tab:ablation_dataset} presents these results, highlighting two key findings:
\begin{enumerate}
\item $\mathbb{SE}(3)$ diffusion consistently demonstrates robust \ac{OOD} detection capabilities across various training datasets;
\item $\mathbb{SE}(3)$ diffusion successfully performs \ac{OOD} detection between two previously unseen datasets during training.
\end{enumerate}
We observe some performance degradation when training with the Oxford Robot Car dataset. This limitation primarily stems from the dataset's restricted trajectory diversity. Both IROS and KITTI datasets exhibit broader data distributions, encompassing more varied trajectory shapes. Consequently, when an Oxford-trained model attempts to distinguish between its own less diverse distribution and a highly varied dataset like IROS, the task becomes particularly challenging. Nevertheless, these results underscore the advantages of our unified diffusion approach to OOD detection. By requiring training on only a single dataset, our method significantly reduces the overall model training time.
\begin{table}[t]
    \caption{AUROC $\uparrow$ of OOD Detection over different train dataset on sequence length of 128 and 30 diffusion steps\\
    (O: Oxford, K: KITTI, I:IROS20)}
    \vskip 0.15in
\begin{adjustbox}{width=0.68\linewidth,center}
    \begin{tabular}{lcccccccc}
    \toprule
        \textbf{Method} & \textbf{O/K} & \textbf{K/O} & \textbf{O/I} & \textbf{I/O} & \textbf{K/I} & \textbf{I/K}\\
    \midrule
        $\mathbb{R}^{3}$-Oxford  & 0.897 & 0.327 & 0.890 & 0.378 & 0.464 & 0.532\\
        $\mathbb{SE}(3)$-Oxford  & 0.934 & \textbf{0.999} & \textbf{1.000} & 0.124 & \textbf{1.000} & 0.433\\
    \midrule
        $\mathbb{R}^{3}$-KITTI  & 0.362 & 0.770 & 0.417 & 0.585 & 0.793 & 0.409\\
        $\mathbb{SE}(3)$-KITTI  & \textbf{1.000} & 0.956 & \textbf{1.000} & \textbf{0.931} & \textbf{1.000} & \textbf{0.897}\\
    \bottomrule
    \end{tabular}
    \end{adjustbox}
    \vskip -0.1in
    \label{tab:ablation_dataset}
\end{table}

\subsubsection{Necessity of Rotational Diffusion Information}
We compare diffusion models trained on translation-only data versus those trained on complete $\mathbb{SE}(3)$ data to demonstrate the critical role of rotational information. Table~\ref{tab:ablation_dataset} reveals that OOD detection using only $\mathbb{R}^{3}$ data yields poor results, consistent with the overlapping statistical distributions shown in figure~\ref{fig:eps_oxford_kitti}. In contrast, $\mathbb{SE}(3)$ diffusion achieves superior performance by incorporating rotational components. This finding demonstrates that for complex trajectory analysis, orientation and rotation data provide richer discriminative features that vary significantly across different data distributions, thereby serving as robust indicators for OOD detection.

\subsubsection{Sequence length of the Trajectory}
Table~\ref{tab:ablation_seqlen} presents the \ac{OOD} detection performance for varying trajectory sequence lengths during KITTI dataset pre-training. The results demonstrate that $\mathbf{DOSE3}$ maintains consistently excellent performance with near-perfect AUROC scores across all \ac{ID} and \ac{OOD} pairs, independent of sequence length. This robustness to sequence length variation highlights the model's stability and generalization capabilities.
\begin{table}[t]
\caption{AUROC $\uparrow$ of OOD Detection over different sequence length for 30 diffusion steps with model trained on KITTI datatset\\
(O: Oxford, K: KITTI, I:IROS20)}
    \vskip 0.15in
    \begin{adjustbox}{width=0.68\linewidth,center}
    \begin{tabular}{lcccccr}
    \toprule
        \textbf{Seq Length} & \textbf{O/K} & \textbf{K/O} & \textbf{O/I} & \textbf{I/O} & \textbf{K/I} & \textbf{I/K}\\
    \midrule
        64   & \textbf{1.000} & \textbf{0.999} & \textbf{1.000} & \textbf{0.986} & 0.999 & \textbf{0.976} \\
        128  & \textbf{1.000} & 0.956 & \textbf{1.000} & 0.931 & \textbf{1.000} & 0.897\\
        256  & 0.980 & 0.961 & \textbf{1.000} & 0.941 & \textbf{1.000} & 0.932\\
        512  & \textbf{1.000} & \textbf{0.999} & \textbf{1.000} & 0.942 & \textbf{1.000} & 0.923\\
    \bottomrule
    \end{tabular}
    \end{adjustbox}
    \vskip -0.1in
    
    \label{tab:ablation_seqlen}
\end{table}

\subsubsection{DDPM Forward Steps}
Table~\ref{tab:ablation_steps} illustrates the relationship between $\mathbf{DOSE3}$ performance and the number of DDPM steps. The results indicate minimal variation in average AUROC scores across different step counts, demonstrating $\mathbf{DOSE3}$'s resilience to changes in the number of DDPM steps.
\begin{table}[t]
\caption{AUROC of OOD Detection over different numbers of diffusion steps on sequence length of 128 with model trained on KITTI datatset\\
(O: Oxford, K: KITTI, I:IROS20)}
    \vskip 0.15in
    \begin{adjustbox}{width=0.68\linewidth,center}
    \begin{tabular}{lcccccr}
    \toprule
        \textbf{Diffusion Step} & \textbf{O/K} & \textbf{K/O} & \textbf{O/I} & \textbf{I/O} & \textbf{K/I} & \textbf{I/K}\\
    \midrule
        5   & 0.916 & 0.969 & \textbf{1.000} & \textbf{0.955} & \textbf{1.000} & \textbf{0.945}\\
        10  & 0.914 & \textbf{0.975} & \textbf{1.000} & 0.938 & \textbf{1.000} & \textbf{0.945}\\
        15  & 0.948 & 0.964 & \textbf{1.000} & 0.919 & \textbf{1.000} & 0.912\\
        30  & \textbf{1.000} & 0.956 & 1.000 & 0.942 & \textbf{1.000} & 0.897\\
    \bottomrule
    \end{tabular}
    \end{adjustbox}
    \vskip -0.1in
    
    \label{tab:ablation_steps}
\end{table}

\section{Conclusions}
Out-of-Distribution (OOD) detection plays a vital role in machine learning, particularly in safety-critical domains like autonomous driving and robotics where systems must reliably interact with the physical world. In these applications, data typically consists of rigid object pose trajectories that capture both positional and rotational motion. While existing OOD detection approaches operate on assumed Euclidean latent spaces, we present $\mathbf{DOSE3}$, a novel unified diffusion-based \ac{OOD} detection framework specifically designed for $\mathbb{SE}(3)$ trajectory data.
$\mathbf{DOSE3}$ innovates by directly incorporating manifold operations into the diffusion model and introduces a novel architecture that extends DDPM to handle $\mathbb{SE}(3)$ manifold sequences. Through comprehensive empirical evaluation across diverse real-world safety-critical datasets, we demonstrate $\mathbf{DOSE3}$'s robust performance and effectiveness.

\bibliography{iclr2025_delta}
\bibliographystyle{iclr2025_delta}

\end{document}